\documentclass[letterpaper, 10 pt, conference]{ieeeconf}
\IEEEoverridecommandlockouts
\usepackage{url}

\usepackage{tabu}
\usepackage{booktabs} % for professional tables
\usepackage{multirow}
\usepackage{multicol}
\usepackage{adjustbox}

\usepackage{xcolor}
\usepackage{xspace}
\usepackage{ifthen}

\usepackage{graphicx}
\usepackage[font=small]{caption}
\usepackage{tikz}
\usetikzlibrary{positioning}

\let\labelindent\relax %needed for IEEE conflict with enumitem
\usepackage{autolabtools}
\usepackage{amsmath,amssymb,amsfonts}
\usepackage{dsfont}
\usepackage{color}
\usepackage{verbatim}
\usepackage{subcaption}
\usepackage{balance}
\usepackage{gensymb}
\usepackage{siunitx} % Daniel: added based on Jeff's prior papers. (needs to be paired with next line :) -Jeff)
\usepackage{algorithm}
\usepackage{algorithmicx}
\usepackage{algpseudocode}
\usepackage{alphalph}
\usepackage{comment}
\usepackage{soul}
\usepackage[shortlabels]{enumitem}

\graphicspath{{figures/}}
\newcommand{\ba}{\mathbf{a}}

\newcommand{\R}{\mathbb{R}}

\newcommand{\mO}{\mathcal{O}}

\algrenewcommand\algorithmicensure{\textbf{Output:}}

\allowdisplaybreaks[4]

\definecolor{britishracinggreen}{rgb}{0.23, 0.53, 0.19}

\definecolor{navy}{rgb}{0,0,0.5}

\newcommand{\bagging}{SLIP-Bagging\xspace}
\newcommand{\slip}{SLIP\xspace}

\usepackage[font={small}]{caption}
\def\tablename{Table}

\usepackage[backend=biber,
            hyperref=true,
            url=false,
            isbn=false,
            doi=false,
            backref=false,
            style=ieee,
            natbib=true,%compatibility aliases
            mincitenames=1,
            maxcitenames=1,
            citestyle=numeric-comp,
            sorting=none,%none vs nyt
            block=none,
            maxbibnames=99]{biblatex}
\usepackage{hyperref}

\title{\LARGE \bf
Bagging by Learning to Singulate Layers Using Interactive Perception
}

\author{Lawrence Yunliang Chen$^1$, Baiyu Shi$^1$, Roy Lin$^1$, Daniel Seita$^2$, Ayah Ahmad$^1$, \\ Richard Cheng$^3$, Thomas Kollar$^3$, David Held$^2$, Ken Goldberg$^1$
\thanks{$^{1}$The AUTOLab at UC Berkeley (autolab.berkeley.edu).}%
\thanks{$^{2}$The Robotics Institute at Carnegie Mellon University.}
\thanks{$^{3}$Toyota Research Institute, Los Altos, USA.}
\thanks{Correspondence to: {\tt\scriptsize yunliang.chen@berkeley.edu}}
}

\begin{document}

\maketitle

\thispagestyle{empty}
\pagestyle{empty}

\begin{abstract}
Many fabric handling and 2D deformable material tasks in homes and industries require singulating layers of material such as opening a bag or arranging garments for sewing. In contrast to methods requiring specialized sensing or end effectors, we use only visual observations with ordinary parallel jaw grippers. We propose \slip: Singulating Layers using Interactive Perception, and apply \slip to the task of autonomous bagging. We develop \bagging, a bagging algorithm that manipulates a plastic or fabric bag from an unstructured state and uses \slip to grasp the top layer of the bag to open it for object insertion. In physical experiments, a YuMi robot achieves a success rate of 67\% to 81\% across bags of a variety of materials, shapes, and sizes, significantly improving in success rate and generality over prior work. Experiments also suggest that \slip can be applied to tasks such as singulating layers of folded cloth and garments.
Supplementary material is available at \url{https://sites.google.com/view/slip-bagging/}.
\end{abstract}

\section{Introduction}\label{sec:intro}
Many tasks in homes and factories require grasping a single layer of 2D deformable objects. Examples include taking one napkin from a stack of napkins, grasping the top layer of a folded towel to unfold it, grasping a single layer of a T-shirt to insert into a hanger, and grasping a single layer of a bag to hold it open while placing items inside. Humans manipulate such deformable objects with great dexterity using touch and vision. 
Such tasks are very challenging for robots. Enabling touch sensing may require equipping the robot end effector with compliant grippers or special tactile sensors such as the mini-Delta gripper~\cite{mannam2021low} and the ReSkin sensor~\cite{bhirangi2021reskin} used in Tirumala~et~al.~\cite{tirumala2022} and GelSight~\cite{yuan2017gelsight} used in Sunil et al.~\cite{sunil2023visuotactile} for cloth manipulation. % The challenge when using visual sensors is occlusion and uncertainty. In industries such as packaging, suction cup arrays and electrostatic mechanisms are a common solution to grasping deformables, but they only work on certain materials (e.g., flat, light-weight, smooth, and without holes). 

%In this work, we do not use any special hardware. 
In this work, we achieve single-layer grasping using a bimanual robot with ordinary parallel-jaw grippers.
We use self-supervised learning to identify where to grasp, and we use interactive perception to determine the number of layers grasped.  The robot iteratively adjusts its grasp until it successfully grasps a single layer.

\begin{figure}[t]
\center
\includegraphics[width=0.49\textwidth]{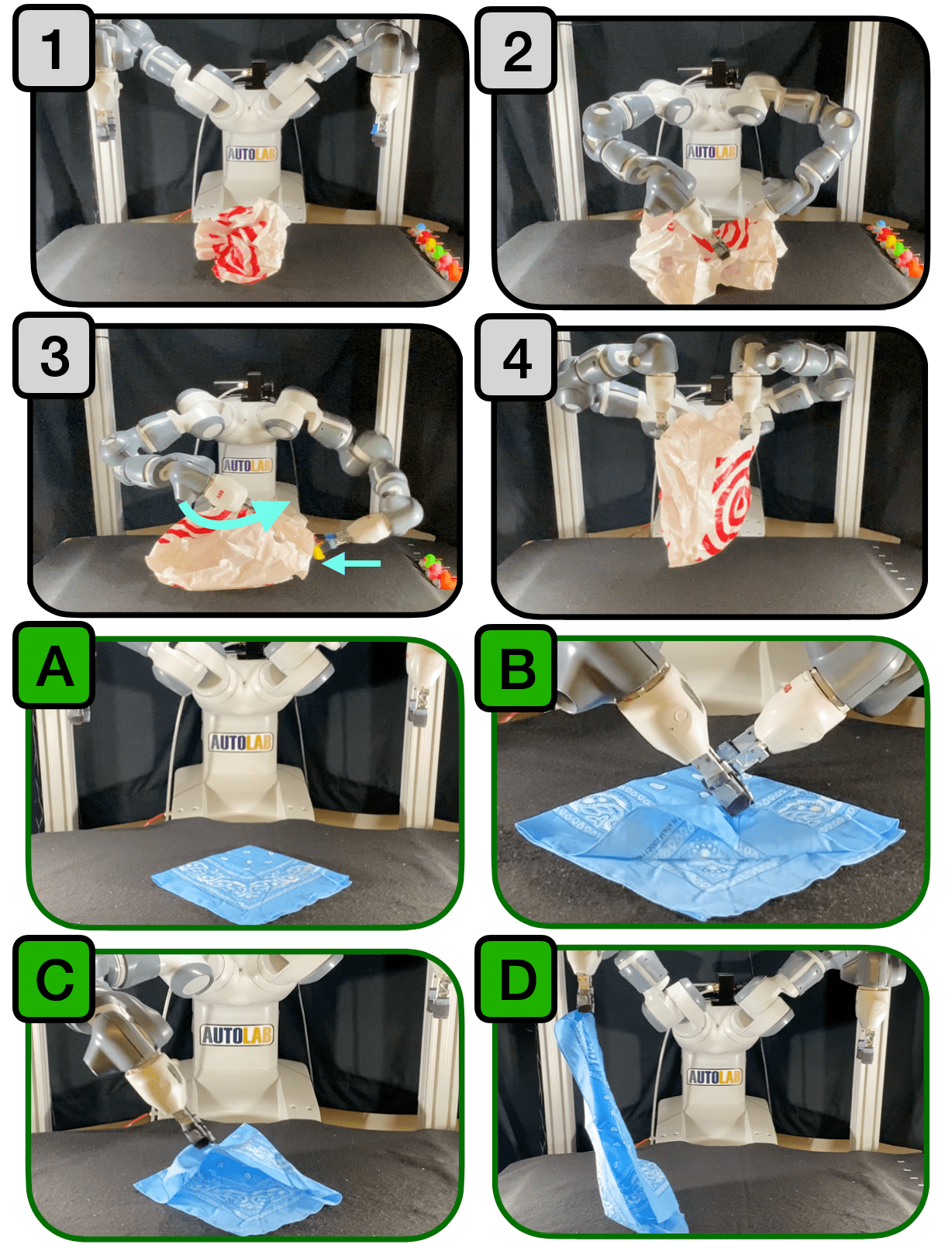}
\caption{
\bagging. \textbf{Top 2 rows:} (1) Initial unstructured and deformed bag. (2) The robot flattens the bag, and then (3) uses \slip to grasp the top layer of the bag. The robot rotates the bag by $90^\circ$ and inserts objects into the bag. (4) The robot lifts the bag filled with the inserted items, so it is ready for transport. \textbf{Bottom 2 rows:} (A) Initial configuration of a piece of folded cloth. (B) The robot uses \slip to grasp the top layer of a folded square cloth. (C) After grasping the top layer, the robot lifts the cloth up. (D) After shaking, the cloth is successfully expanded.
}
\vspace{-15pt}
\label{fig:teaser}
\end{figure}

We focus on the task of autonomous bagging --- opening a deformable bag from an unstructured initial state and putting objects into it. This task has wide applications in retail, food handling, home cleaning, and packing. However, it involves perception and manipulation challenges for robots. %Compared to ropes and fabrics, there are more ways for 3D bags to deform and create occlusion. 
There are many ways for bags to deform and create self-occlusions.
Many plastic bags are also reflective and translucent, as well as elastoplastic, meaning they have a tendency to restore their shape under small forces. However, an ideal state for object placement---a bag standing upward with its opening wide open---is also not a naturally stable pose for soft deformable bags, which tend to lie on their side. An alternative approach is to open the bag and insert objects horizontally, but this requires grasping only the top layer to create the bag opening. This is nontrivial because: (1) The depth values from a typical RGBD camera are often severely inaccurate on bags due to reflection and transparency; (2) even if the depth of the top layer is known, moving the gripper to that height to perform the grasp will push the surface downward and cause a missed grasp due to lack of friction on the surface; and (3) grasping from the side is not always kinematically feasible since the two layers can be stuck together with no space in between.
%; and (4) the only reliable way to grasp the top layer from the top is to push it down to the table or other support surfaces and use the friction to pinch the bag while closing the gripper, but this often also accidentally grasps the bottom layer. 
As we demonstrate in experiments, a 1 mm change in the gripper height can lead to the difference between a missed (0-layer) grasp, a 1-layer grasp, and a 2-layer grasp of plastic bags, and the heights of successful 1-layer grasps are different each time depending on the shape of the bag (wrinkles and flatness of both the top and bottom layers).

We use interactive perception~\cite{bohg2017} to recognize how many layers a robot grasps, and to adjust its grasp if needed. While the robot cannot a priori know what grasp height it should go to grasp only one layer of the bag, after a grasp is performed it can tell how many layers it has grasped by perturbing the bag and observing how the bag moves with the gripper. Intuitively, if the bag does not move, it indicates a 0-layer grasp. If the top layer of the bag moves with the gripper while the bottom layer only moves a little and mostly stays on the table, it indicates a 1-layer grasp. If the entire bag moves with the gripper, it suggests a 2-layer grasp. 

We propose \slip: Singulating Layers using Interactive Perception. Using \slip, we present an algorithm for opening deformable bags, which we call \bagging, that is effective across a variety of bag materials, including non-reusable plastic bags such as thin and soft bags made of low-density polyethylene (LDPE) and thicker and stiffer grocery bags made of high-density polyethylene (HDPE), as well as reusable fabric bags such as drawstring backpacks, mesh drawstring bags, and large fabric handbags. Physical experiments suggest that \bagging achieves a 5x success rate compared to a prior state-of-the-art method for autonomous bagging~\cite{chen2022autobag}. Moreover, we conduct physical experiments to evaluate the applicability of \slip to singulating layers for a variety of fabrics and garments (see Fig.~\ref{fig:teaser}).

This paper makes the following contributions:
\begin{enumerate}[leftmargin=*]
    \item \slip, an algorithm to singluate layers of bags using interactive perception, with visual feedback to enable the robot to adapt its grasp height without tactile sensors;
    \item \bagging, an algorithm that uses \slip for opening and inserting objects into a deformable bag in an unstructured initial state;
    \item A \bagging system and physical experiments that achieve a success rate of 67\% to 81\% across bags of different materials, shapes, and sizes (unseen in training). On thin plastic bags, \bagging's success rate is 5x that of the state-of-the-art method designed specifically for thin plastic bags. 
    \item Physical experiments suggesting the applicability of \slip to other tasks such as singulating layers of fabrics and dresses.
\end{enumerate}

\section{Related Work}

\subsection{Deformable Objects and Single-Layer Grasping}
There is a rich literature on deformable object manipulation; see~\cite{manip_deformable_survey_2018,grasp_centered_survey_2019,2021_survey_defs} for representative surveys. Deformable objects are challenging due to their infinite degrees of freedom, which both induce complex dynamics that are hard to model and control and lead to self-occlusions that makes planning challenging. 
%\daniel{I'd cut at least the 1D stuff out} For 1D deformable objects such as cables and ropes, prior work has studied tasks such as insertion~\cite{wire_insertion_1996,wire_insertion_1997}, knot tying and untangling~\cite{knot_planning_2003,tying_precisely_2016, untanglingLongCables2022,grannen2020untangling,SundaresanGrannen-RSS-21}, and reaching goals or target configurations~\cite{zhu_sliding_cables_2019,nair_rope_2017,wang_visual_planning_2019, harry_rope_2021,chi2022irp,lim2022real2sim2real}. She et al.~\cite{tactile_cable_2020} use a tactile-reactive gripper with a GelSight sensor to perform cable following and insertion tasks. 
Among deformable object manipulation, fabric manipulation is one of the most widely-studied areas~\cite{flinging_2022,fabric_vsf_2020,cloth_region_segmentation_2020,seita-bedmaking,ha2021flingbot,fabricflownet,VCD_cloth,lerrel_2020,gdoom2021,speedfolding_2022}. %bodies_uncovered_2022, --cut for space
These works focus on learning grasp locations that are effective for pick-and-place actions or dynamic actions to achieve smoothing and folding for one piece of fabric.

% Daniel: commenting the below out as these probably are not good fits; the fabrics one we can argue from a layering or a 2D perspective. And since it's IROS I think we have to get all references within the 8 page limit, unfortunately, so we should add more from interactive perception (I think Shuran Song also has papers in that area).
%Ichiwara et al.~\cite{ichiwara2022contact} use tactile sensors to perform the unzipping task. Beyond fabrics, prior work has also studied manipulating plush toys, sponges, and dough~\cite{ACID2022,Qi_dough_2022,matl2021Deformable,PASTA_2022}, or objects typically held in containers, such as liquids~\cite{visual_closed_loop_liquids_2017} and granular media~\cite{schenck_2017,samuel_clarke_2018,matl2020inferring}.

While some prior work has studied singulating a single sheet or fabric layer from a stack, most use tactile sensing or specialized end effectors.  Tirumala~et~al.~\cite{tirumala2022} use a ReSkin sensor~\cite{bhirangi2021reskin} to singulate layers of cloth from tactile feedback. Manabe~et~al.~\cite{manabe2021} design a rolling hand mechanism to separate a single sheet from a pile of fabrics. Guo~et~al.~\cite{deformation_page_turning_2021} use a XELA uSkin tactile sensor combined with visual inputs to turn a single book page. In this work, we propose to singulate layers with standard end effectors purely from visual feedback. Demura~et~al.~\cite{demura2018} study grasping the top folded towel from a stack using visual feedback with a scooping action, using towels which are each several millimeters thick. In contrast, we study manipulation tasks where layers can be thinner than 1 mm. 

\subsection{Manipulating Deformable Bags}
Some early work on bag manipulation studies mechanical design or policies for grasping~\cite{grasping_sacks_2005}, lifting~\cite{ayanna_2000} or unloading~\cite{unloading_sacks_2008} large sacks. 
%Recent work also studies more complex contact-rich bag manipulation tasks such as closing ziplock bags~\cite{contour_ziplock_2018}, unzipping a fabric bag~\cite{ichiwara2022contact}, and tying the handles of a plastic bag~\cite{bagKnotting2022}. 
Prior work also studies bag manipulation in simulation. For example, Seita~et~al.~\cite{seita_bags_2021} benchmark several simulation tasks that involve opening a bag and inserting objects into it, and Weng~et~al.~\cite{graph_based_interaction_2021} study modeling bags using graph neural networks. Much of the prior work on physical experiments with deformable bags assume a semi-structured bag state, such as pregrasped~\cite{contour_ziplock_2018,ichiwara2022contact,dextairity2022}, filled with objects~\cite{bagKnotting2022},  oriented upwards with the bag wide open~\cite{seita_bags_iros_2021,bahety2022bag}, and focus on a specific task such as packing and arranging objects~\cite{bahety2022bag}, opening the bag~\cite{dextairity2022}, or lifting the bag~\cite{seita_bags_iros_2021}. In contrast to these works, we study physical bag manipulation where bags start in unstructured states.

% Daniel (March 01): made some cuts here since this was too long.
Recently, Chen~et~al.~\cite{chen2022autobag} propose the AutoBag algorithm for manipulating a thin plastic bag from an unstructured state. In their setting, the bags can be compressed, deformed, and arbitrarily oriented, and the task is to reorient the bag upward, enlarge the opening, insert objects, and then lift the bag up. 
%For perception, Chen~et~al. propose to represent bags using key semantic parts (\eg handles and rim) learned with self-supervision from UV-fluorescent markings. For manipulation, they propose several novel primitives such as ``Compress,'' ``Dilate,'' ``Flip,'' and ``Pin-Pull.'' 
%However, their success rate is low (1/6) 
However, AutoBag frequently fails when attempting to orient the bag upward. %since it is not a stable pose for deformable bags. 
Gu et al.~\cite{gu2023ShakingBot} improve AutoBag by using dynamic shaking actions and performing item insertion with one gripper grasping the bag handle in midair.
% In contrast to AutoBag, we 
%use AutoBag's perception representation but 
% avoid orienting the bag upward. Instead, 
In contrast, we flatten the bag, singulate the top layer to open it, and insert objects sideways, which results in much higher success rates. Moreover, while AutoBag is designed specifically for opening thin plastic bags, we show evidence that \bagging is effective on other bag materials and shapes. %since flattening is easier than orienting a bag upward, and fabric smoothing methods can be applied to flattening a fabric bag.

\subsection{Interactive Perception}

%Interactive perception integrates perception and action, allowing robots to actively gather information about the environment through physical interactions with objects. 
%This approach recognizes that perception and action are tightly intertwined and that a robot's ability to perceive its environment can be greatly enhanced through active exploration and manipulation. 
Interactive perception combines perception and action, enabling a robot to reduce uncertainty through active physical exploration and interaction with objects~\cite{bohg2017}.
Interactive perception has a variety of applications, ranging from manipulation to object segmentation, to grasp planning~\cite{IP_cluttered_2020}. 
Recent work has used interactive perception to better understand properties of geometrically challenging objects for robotic manipulation. For example,~\cite{gadre2021act,nie2022sfa} study how to interact with articulated objects to discover their geometric information.
For deformable object manipulation, Shivakumar~et~al.~\cite{untanglingLongCables2022} use interactive perception to autonomously untangle cables, and Willimon~et~al.~\cite{Willimon2011} use interactive perception to detect and classify clothing. In this work, we use interactive perception to identify and singulate individual layers of bags to improve robotic bag manipulation.

\section{Problem Statement}\label{sec:PS}

% Daniel: rearranged here as we introduce the bags here.
\begin{figure}[t]
\center
\includegraphics[width=0.42\textwidth]{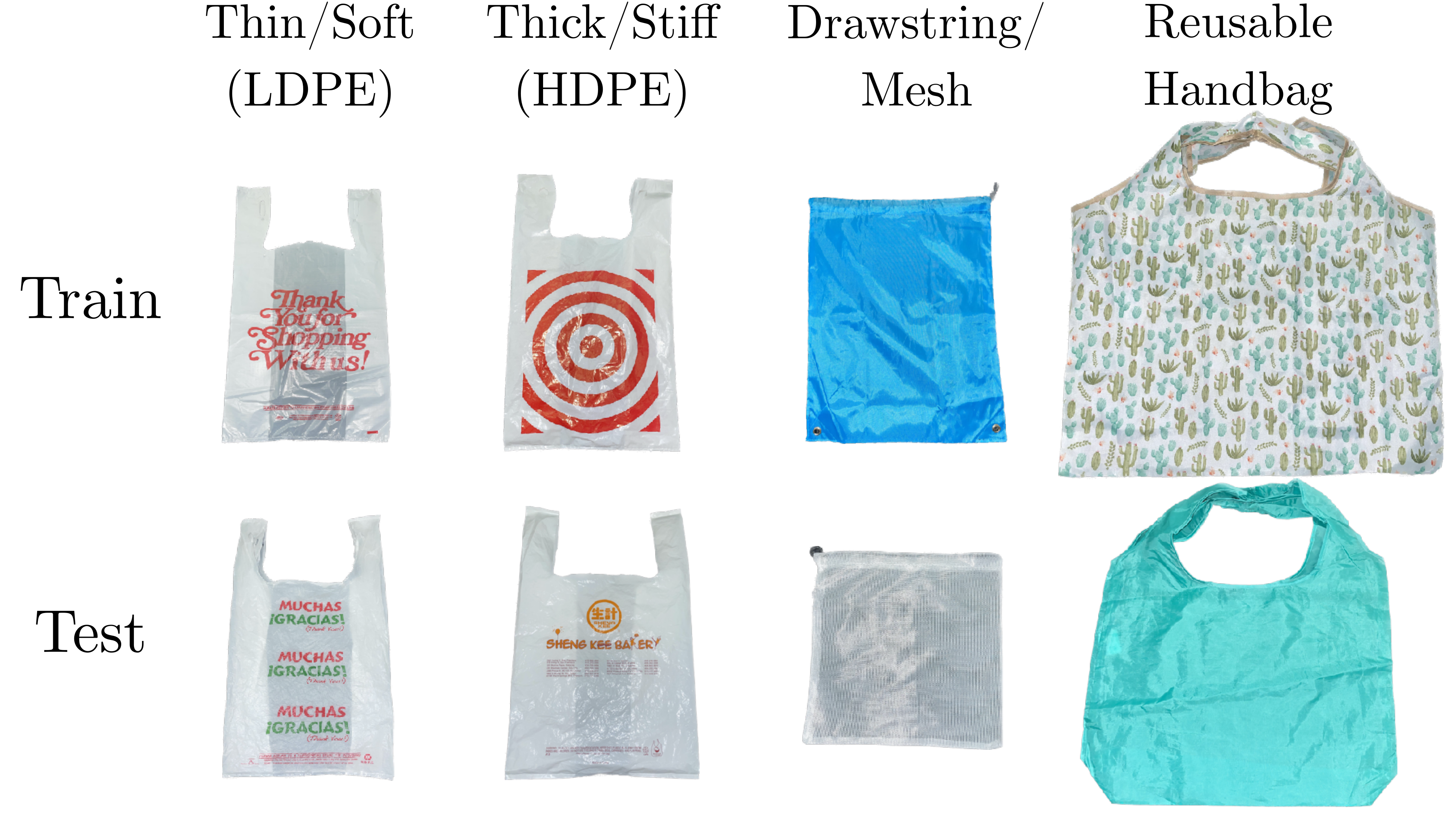}
\caption{
Top: Training bags. Bottom: Test bags. See Section~\ref{sec:exp} for more details and Table~\ref{tab:bag_result} for results.
% Categories from left to right: thin LDPE plastic bags, thick HDPE plastic bags, drawstring bags, and reusable handbags.
}
\vspace{-15pt}
\label{fig:bags}
\end{figure}

We study the autonomous bagging task. As defined in prior work~\cite{chen2022autobag}, the task consists of manipulating and opening a deformable bag from an unstructured state, inserting $n$ items, and then lifting it for transport. Unlike prior work~\cite{chen2022autobag} which only considers thin plastic bags made of LDPE, we additionally test heavy-duty grocery plastic bags made of HDPE, drawstring backpacks and mesh bags, and reusable fabric handbags (as shown in Figure~\ref{fig:bags}).

We consider a bimanual robot with two standard parallel jaw grippers, operating in an $(x, y, z)$ cartesian coordinate frame with a flat manipulation surface parallel to the $xy-$plane and a calibrated overhead RGBD camera. At each time step, the robot uses the RGBD input $I \in \R ^ {W \times H \times 4}$ of the bag to select and execute a parameterized open-loop action primitive $\mathbf{a}$ according to a policy $\pi: I \mapsto \ba$. We assume the initial bag state is unstructured and resting stably on the workspace, that is, it may be deformed and compressed, but not tied or zipped. We assume a set of rigid objects $\mO$, placed in known poses for grasping. We use the following metrics for the task: (1) the success rate of grasping a single layer, (2) the percentage of objects successfully inserted into the bag, and (3) the percentage of objects contained in the bag after the robot lifts the bag.

We make the following assumptions: (1) the two sides of the bag do not stick to each other tightly (which occurs in brand-new plastic bags), (2) the bag can be segmented from the workspace via color thresholding, and (3) the size of the bag when fully flattened, denoted $A_{max}$, is known a priori.

\section{\slip: Singulating Layers using Interactive Perception}\label{sec:slip}

% Daniel: I edited Lawrence's slides for slip_v2. Also v3.
% Also moved this figure here in the section that describes SLIP.
\begin{figure*}[t]
\center
\includegraphics[width=1.00\textwidth]{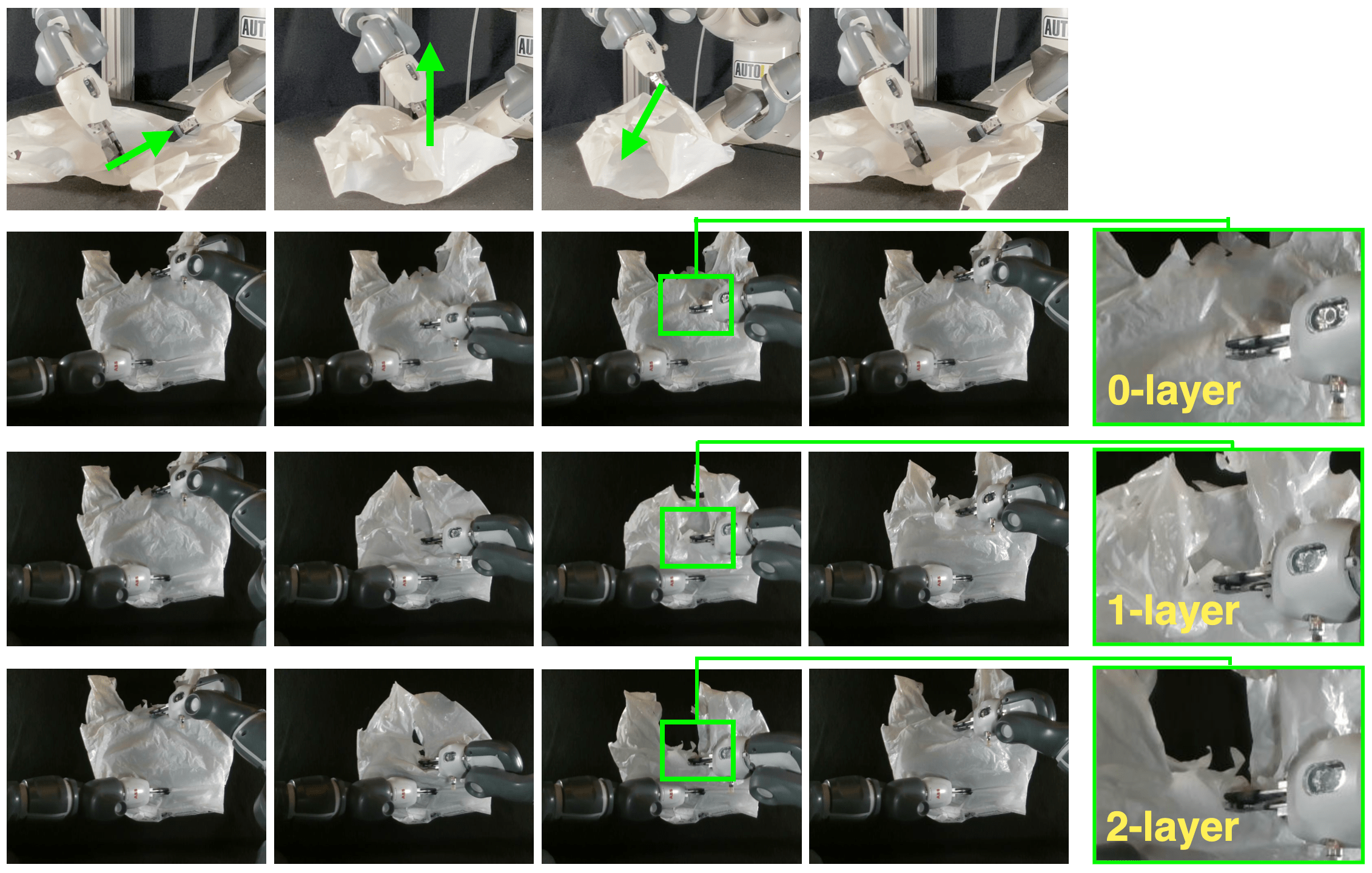}
\caption{
Examples of \slip in action. Each row shows an example of one cyclic, triangular trajectory $T$ in an iteration (``iter'' in Algorithm~\ref{alg:slip}) where one gripper moves the bag while the other one pins the bag. \textbf{Top row} a third-person view of the robot \textbf{Next three rows}: top-down RGB camera views of one cyclic trajectory for different trials. They show, respectively, a 0-layer, 1-layer, and 2-layer grasp on a plastic bag. We provide zoomed-in versions of images in the third column to see the layers in more detail. See Section~\ref{sec:slip} for details.
}
\vspace{-10pt}
\label{fig:slip}
\end{figure*}

In this section, we describe \slip in the context of grasping a single layer of a deformable bag, but the algorithm may apply to other fabric materials (see Section~\ref{sec:exp}). As discussed in Section~\ref{sec:intro}, \slip is motivated by how, after the robot has performed a grasp, it cannot easily determine how many layers it grasped from visual inputs of a static scene, as the top layer is occluding the layers underneath. However, by moving the gripper and observing how the bag is moved with the top layer, the robot can infer how many layers it has grasped. Formally, \slip requires 3 components: a cyclic trajectory $T$ of the robot gripper, a video classification model $M$, and an iterative height adjustment algorithm. %We describe each component in detail next.

\begin{algorithm}[t]
\caption{\slip: Singulating Layers using Interactive Perception}\label{alg:slip}
% \footnotesize
\begin{algorithmic}[1]
\Require RGBD camera, closed trajectory $T$ of the gripper, video classification model $M$, initial grasp height $h_0$, iterative height adjustment $\Delta h_+, \Delta h_-$, maximum number of iterations TrialMax, IterMax
\Ensure Single-layer grasp success
% \Statex
\State Grasp height $h = h_0$, trial = 0
\While{trial $<$ TrialMax}
    \State Sample a grasp location $(x, y)$
    \State iter = 0
    \While{iter $<$ IterMax}
        \State Grasp at location $(x, y, h)$
        \State Execute trajectory $T$ and record a video stream $V$ from the camera during the motion
        \State Number of layers grasped $n = M.predict(V)$
        \If{$n = 0$}
            $h \gets h - \Delta h_-$
        \ElsIf{$n = 1$}
            return True
        \ElsIf{$n \geq 2$}
            $h \gets h + \Delta h_+$
        % \ELSE
        %     \State success = True
        \EndIf
        \State Open gripper
        \State iter $\gets$ iter $+1$ 
    \EndWhile
        \State trial $\gets$ trial $+1$
\EndWhile
\State return False
\end{algorithmic}
\end{algorithm}

The trajectory $T$ needs to satisfy two properties: (1) The movement should reveal enough information for the robot camera to infer how many layers are grasped, and (2) the trajectory should be cyclic, so the bag roughly recovers its original state after executing $T$, allowing the robot to retry the grasp at the same location but with a different height. For (1), the main consideration is occlusion, as the robot gripper and wrist occlude the grasp point and its nearby region if we use a top-down grasp with an overhead camera. Thus, we tilt the gripper at an angle $\theta = 50^{\circ}$ so the grasp point is visible in the camera. For (2), we use a triangular trajectory, where the robot gripper first moves backward, then upward, and finally forward and downward back to the original position. To prevent the deformable object from translating as a whole, we use the robot's second gripper to pin the other side. See Figure~\ref{fig:slip} for a visualization. In our implementation, the trajectory takes about 5 secs.

While the robot executes the trajectory $T$, the camera takes an RGB video stream of the bag. A video classification model $M$ takes the video and classifies the grasp into 3 categories: 0 layer, 1 layer, and 2 layers. We use a SlowFast network with a ResNet-50 backbone~\cite{feichtenhofer2019slowfast}, which takes in 32 images of size 224 $\times$ 224 sampled with a uniform interval from the video stream. Fig.~\ref{fig:slip} illustrates the visual differences among different layers grasped on a plastic bag.

Given the model classification, \slip adjusts the gripper height and retries the grasp if it does not successfully grasp a single layer. We choose to use a fixed height adjustment each time which is similar to the strategy in~\cite{tirumala2022}, with height deltas $\Delta h_{-}$ and $\Delta h_{+}$. One could also choose to let the adjustment height decay over time or use the bisection method, but we empirically find the numbers of trials these approaches take are similar while a fixed height adjustment is more robust to a long sequence of iterations than a decaying step size, and more robust to minor bag state changes between grasps and model classification errors than bisection, since once bisection enters into a wrong interval, it may not succeed. The pseudocode of \slip is provided in Algorithm~\ref{alg:slip}.

\section{\bagging}

% Daniel: modified Lawrence's keynote to get this figure.
% Also put this by the section which actually describes SLIP-Bagging.
\begin{figure*}[t]
\center
\includegraphics[width=1.00\textwidth]{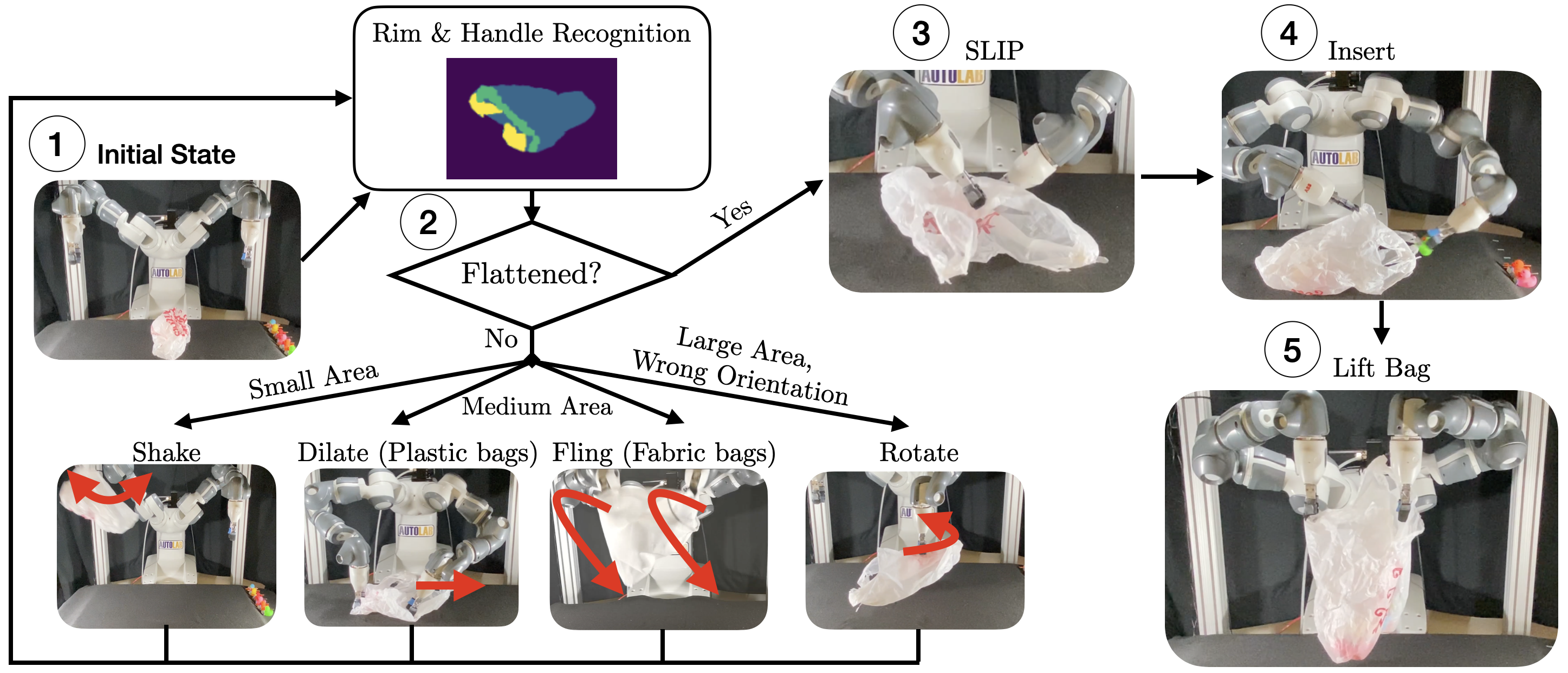}
\caption{
\bagging Algorithm. (1) The robot starts with an unstructured bag with objects on the side. (2) \bagging then flattens the bag, and (3) uses \slip to grasp the top layer of the bag, followed by (4) insertion and (5) bag lifting. A trial is a full success if the robot lifts the bag with all items in it.
}
\vspace{-10pt}
\label{fig:slip-bag}
\end{figure*}

\subsection{Learned Perception Module}\label{subsec:perception_module}
Similar to Chen et al.~\cite{chen2022autobag}, we train a perception module to recognize the bag rim and handles. We represent bags through semantic segmentation, where each RGB image is classified per pixel into: bag handle, bag rim, remaining area of the bag, and the background. To collect training data, we use self-supervision with UV-fluorescent markings~\cite{LUV_2022}. We place 6 UV LED lights around the workspace and paint the handles and rim of the bags with UV-fluorescent markers. During data collection, we take image pairs of the bag under regular lighting and under UV lighting. When the UV lights are turned on, the painted regions glow unique colors that can be extracted in the image through color thresholding, allowing us to get ground truth segmentation labels corresponding to the bag image under regular lighting. The robot then performs a random action to disturb the bag into another state, and repeats the process. This allows us to obtain a large dataset with diverse bag configurations efficiently without the need of human annotations. See the project website and~\cite{chen2022autobag} and~\cite{LUV_2022} for details and examples. 
%The main difference in this work compared to AutoBag~\cite{chen2022autobag} is that we show how this procedure applies to bags of other material and shapes such as a fabric bag (see Figure~\ref{fig:bags}).

\subsection{Action Primitives}
Following~\cite{chen2022autobag}, \bagging uses 4 manipulation primitives to flatten a bag from an unstructured state:
\begin{enumerate}[leftmargin=*]
\item \textbf{Shake}: Grasp a corner of the bag, lift it up, shake it a predefined number of times, followed by a swing action to lay the bag on the table. This action uses gravity and inertia to loosen the bag.
\item \textbf{Rotate}: For small plastic bags, the robot uses one hand to rotate them. It grasps the center of the bag and rotate a desired angle. For large fabric bags, the robot uses two hands since one hand is not effective due to underactuation. It grasps the left and right sides of the bag and pushes one hand forward while pulling the other hand backward to rotate the bag.
\item \textbf{Dilate}: First, the left gripper pins the left side of the bag, while the right gripper presses the bag and moves from the center to the right. Then, the right gripper pins the right side of the bag, while the left gripper presses the bag and moves from the center to the left. This action flattens the bag.
\item \textbf{Fling}: Inspired by prior work that uses a fling action to smooth garments~\cite{ha2021flingbot,flinging_2022,speedfolding_2022}, we design a fling primitive to smooth fabric bags. Given two pick points, the robot lifts the bag above the surface with two hands and flings the bag forward and then backward while putting it down. The fling velocity is set to the robot's maximum speed.
\end{enumerate}
In addition, the robot recenters the bag as needed. For \textbf{Dilate}, the robot always first rotates the bag so that the bag's shorter axis aligns with the robot's horizontal axis, as \textbf{Dilate} can help expand that axis using friction. See the website for videos.

\subsection{\bagging}
The \bagging algorithm consists of 5 steps: (1) flatten the bag, (2) grasp the top layer of the bag near the bag opening using \slip, (3) rotate the bag sideways, (4) use the other gripper to insert objects, and (5) lift the bag. See Figure~\ref{fig:slip-bag} for an overview.

\begin{figure*}[t]
\center
\includegraphics[width=1.00\textwidth]{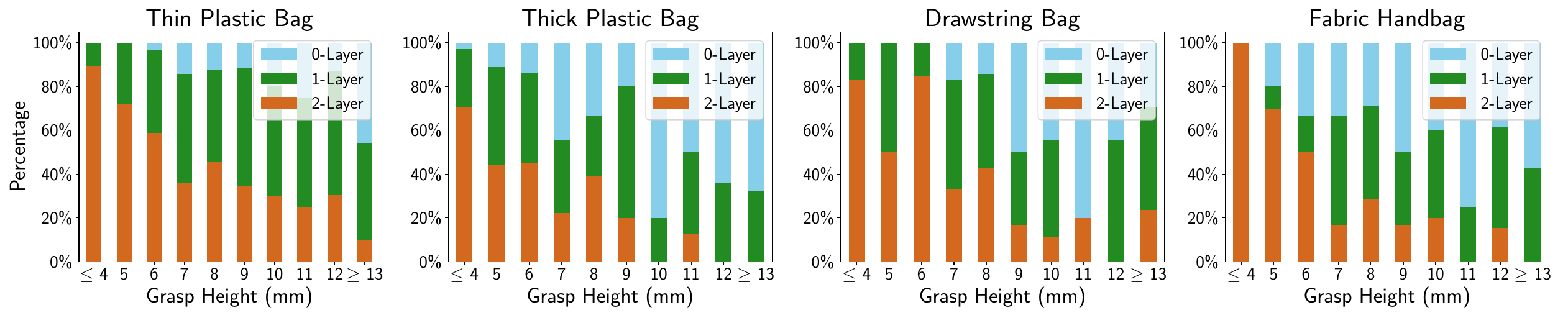}
\caption{
Distribution of the number of layers grasped for different grasp heights for 4 different bags.
}
\vspace{-5pt}
\label{fig:grasp_height}
\end{figure*}

\begin{table*}
\centering
\begin{tabular}{cc|cc| ccc| cccc| cccc | c}
    \toprule
\multirow{2}{*}{Category} 
 & \multirow{2}{*}{Bag}   & \multicolumn{2}{c}{Open/Flatten} & \multicolumn{3}{c}{Single-layer Grasp}  & \multicolumn{4}{c}{Full Success} & \multicolumn{4}{c}{\% Objects Inserted} & \multirow{2}{*}{\parbox{1.3cm}{\centering SB Failure Modes}} \\
    \cmidrule(lr){3-4} \cmidrule(lr){5-7} \cmidrule(lr){8-11} \cmidrule(lr){12-15} 
        &  &  AB  & SB  & PD  & HG   & SB  & PD  & HG  & AB  & SB & PD  & HG  & AB  & SB        \\
    \midrule
    \midrule
\multirow{2}{*}{Thin Plastic}  
        & Train    & 3/6    & \textbf{6/6}    & 1/6    & -    & \textbf{6/6}   & 1/6    & 0/6    & 1/6    & \textbf{5/6}   & 17\%   & 0\%    & 39\%   & \textbf{94\%}   &  (D)     \\
         & Test    & 3/6    & \textbf{6/6}    & 1/6    & -    & \textbf{6/6}   & 1/6    & 0/6    & 1/6  & \textbf{4/6}   & 17\%   & 0\%    & 36\%   & \textbf{75\%}   &  (D) $\times$ 2    \\
    \addlinespace
    \midrule
\multirow{2}{*}{Thick Plastic}
        & Train    & 3/6    & \textbf{5/6}    & 1/6     & -    & \textbf{5/5}$^*$   & 1/6    & 0/6    & 1/6 & \textbf{3/6}   & 17\%    & 0\%    & 36\%   & \textbf{56\%}   & (A) (C) (D)      \\
         & Test    & 2/6    & \textbf{5/6}    & 0/6    & -    & \textbf{4/5}$^*$  & 0/6    & 0/6    & 2/6 & \textbf{4/6}   & 0\%    & 0\%    & 33\%   & \textbf{67\%}   & (A) (B)     \\
    \addlinespace
    \midrule
\multirow{2}{*}{Drawswtring}
         & Train    & 0/6    & \textbf{6/6}    & 0/6    & -   & \textbf{5/6}   & 0/6   & -    & 0/6 & \textbf{5/6}   & 0\%   & -    & 0\%   & \textbf{83\%}   &  (B)     \\
         & Test    & 0/6    & \textbf{6/6}    & 1/6    & -   & \textbf{5/6}   & 0/6    & -    & 0/6 & \textbf{4/6}   & 0\%    & -    & 0\%   & \textbf{67\%}   &  (B) (C)    \\
    \addlinespace
    \midrule
% \multirow{2}{*}{Reusable}
Reusable        & Train    & 0/6    & \textbf{5/6}    & 0/6    & 5/6    & \textbf{3/5}$^*$   & 0/6    & 2/6   & 0/6 & \textbf{3/6}   & 0\%    & 36\%    & 0\%   &  \textbf{50\%}   &   (A) (B) $\times$ 2    \\
Handbag         & Test    & 0/6    & \textbf{6/6}    & 1/6   & 4/6    & \textbf{6/6}  & 1/6    & 3/6    & 0/6 & \textbf{4/6}   & 17\%    & 50\%    & 0\%   & \textbf{81\%}   &  (C) (D)      \\
    \bottomrule \\
\end{tabular}
$^*$Denominator is the number of successful flattened trials that proceed to the \slip stage. 
\caption{Physical experiment results of \bagging compared with baselines. 6 trials were run on each of the 8 bags (Fig.~\ref{fig:bags}) for each method. Each trial attempts to insert 6 rubber ducks, and ``\% Objects Inserted'' is the average percentage of ducks inserted and remaining in the bag after bag lifting. \textbf{PD}: Perceived-Depth baseline. \textbf{HG}: Handle Grasping baseline. \textbf{AB}: AutoBag. \textbf{SB}: SLIP-Bagging. The ``Single-layer Grasp'' column for HG refers to grasping the handle associated with the top layer for handbags whose handles overlap (and thus is not applicable to plastic bags and drawstring bags). See Section~\ref{subsec:bag_results} for failure mode categories.
}
\label{tab:bag_result}
\vspace*{-10pt}
\end{table*}

In the first step, the robot iteratively checks the bag area and orientation at each time step to determine whether the bag reaches a ``flattened'' state ready for single-layer grasping. \bagging requires 3 threshold hyperparameters, $p_\text{small}$, $p_\text{large}$, and $\alpha$. For a bag with an area $A_{max}$ when fully flattened, if the current bag area is below $p_\text{small}A_{max}$, the robot performs the \textbf{Shake} primitive. If the current bag area is between $p_\text{small}A_{max}$ and $p_\text{large}A_{max}$, the robot uses \textbf{Dilate} to expand a plastic bag and \textbf{Fling} to expand a fabric bag. If the current bag area is greater than $p_\text{large}A_{max}$, the robot checks whether the bag opening is pointing forward. If the angle between the bag's opening and the robot's forward axis is greater than $\alpha$, the robot \textbf{Rotates} the bag; otherwise, the bag is considered successfully ``flattened.'' We set $\alpha = \pi/8$ and please see project website for choosing $p_\text{small}$ and $p_\text{large}$. 

Next, the robot uses \slip to grasp the top layer of the bag. One hand pins the bottom of the bag while the other hand grasps a point near the center of the rim. We set the initial height of the grasp $h_0 = \max(h_{PD}, h_{min})$, where $h_{PD}$ is the perceived depth of the grasp point on the bag surface measured by the RGBD camera, and $h_{min}$ is a threshold to prevent the robot to grasp too deep in the case of erroneous depth measurements such as for mesh bags that have holes. We set $\Delta h_{-} = 1$ mm and $\Delta h_{+} = 3$ mm.

After successfully grasping the top layer of the bag, the robot rotates the bag by $90^\circ$ while the other arm grasps the inserted items and performs sideways insertion. Finally, the other arm goes inside the bag, grasps it, and lifts the bag together with the arm that has been holding the bag.

\section{Physical Experiments}\label{sec:exp}

For experiments, we use a bimanual ABB YuMi robot with an overhead RealSense D435 camera. The workspace has dimensions 60$\times$90 cm$^2$, and the bags we use range from 32$\times$32 cm$^2$ to 55$\times$55 cm$^2$.

\subsection{Implementation Details}
To train the perception module to recognize the rim and handles of bags, we collect a total of 7,500 images across 15 bags in 4 categories, with about 500 images for each bag, using the self-supervised process described in Sec.~\ref{subsec:perception_module}. The 4 categories are: non-reusable thin plastic bags made of LDPE, non-reusable thick plastic bags made of HDPE, drawstring bags including backpacks and mesh bags, and reusable fabric handbags. For the segmentation network, we use a U-Net architecture~\cite{ronneberger2015u} trained with soft DICE loss~\cite{milletari2016v}. We use one NVIDIA A100 GPU, with a batch size of 32, an initial learning rate of 5e-4, and a weight decay factor of 1e-5. The trained model achieves a 70\% intersection over union (IOU) on the validation set. 

To train the video classification model for \slip, we collect 800 video examples on 4 bags (one in each category), with 200 each. During data collection, for each sample, we manually set the bag into a roughly flat state, and then specify a grasp point as well as a grasp height. We randomize the grasp height so that the number of 0-layer, 1-layer, and 2-layer examples are balanced in the dataset. 
As described in Sec.~\ref{sec:slip}, we use a SlowFast network architecture~\cite{feichtenhofer2019slowfast} with cross-entropy loss. 
%The 32 image frames the model takes in are sampled from uniform intervals. 
We train the model on an A100 GPU with a batch size of 32, a learning rate of 5e-4, and an Adam optimizer. The trained model achieves a 90\% accuracy on the validation set.

% Daniel: this reference was, bafflingly, cited when we wrote the 'ducks'!
% ~\cite{Burgert2022triton}
\subsection{Bagging Experiments Setup}
We evaluate \bagging on 8 bags, shown in Figure~\ref{fig:bags}. We use 6 rubber ducks of dimension $6 \times 7 \times 5$ cm$^3$ as the objects for insertion. For each trial, we randomly initialize the bag state by taking the bag, compressing and deforming it with our hands, dropping it onto the workspace, and letting the bag settle into a stable state. We allow for up to 30 actions (excluding recentering) for the robot to flatten the bag and up to 15 grasp iterations during \slip. If the robot encounters motion planning or kinematic errors during the trial, we reset the robot to the home position and continue the trial.

We compare \bagging to 3 baselines:
\begin{enumerate}[leftmargin=*]
    \item Perceived-Depth (PD): This method ablates the \slip algorithm in \bagging. Instead, the robot directly grasps at $h_{PD}$, the perceived depth of the grasp point on the bag surface measured by a depth camera. %Unlike \bagging, PD requires a depth camera (in addition to RGB information).
    \item Handle Grasping (HG): This method selects a handle to grasp instead of the bag rim. After grasping the handle and lifting the bag up, the robot performs sideways insertion underneath the grasping hand, similar to \bagging.
    \item AutoBag (AB)~\cite{chen2022autobag}: AutoBag uses ``Compress'' and ``Flip'' primitives to orient the bag upward and open the bag directly. It places the objects from the top onto the bag opening and then lifts the bag up instead of inserting the objects sideways.
\end{enumerate} 
For Perceived-Depth and Handle Grasping, we evaluate from an already-flattened bag state since the procedure of flattening the bag is the same as that in \bagging. For \bagging, we record the success rate of flattening the bag and grasping a single layer (which opens the bag from the side). For AutoBag, we measure the success rate of opening the bag upward. For each bag, we conduct 6 trials. 

\begin{figure}[t]
\center
\includegraphics[width=0.3\textwidth]{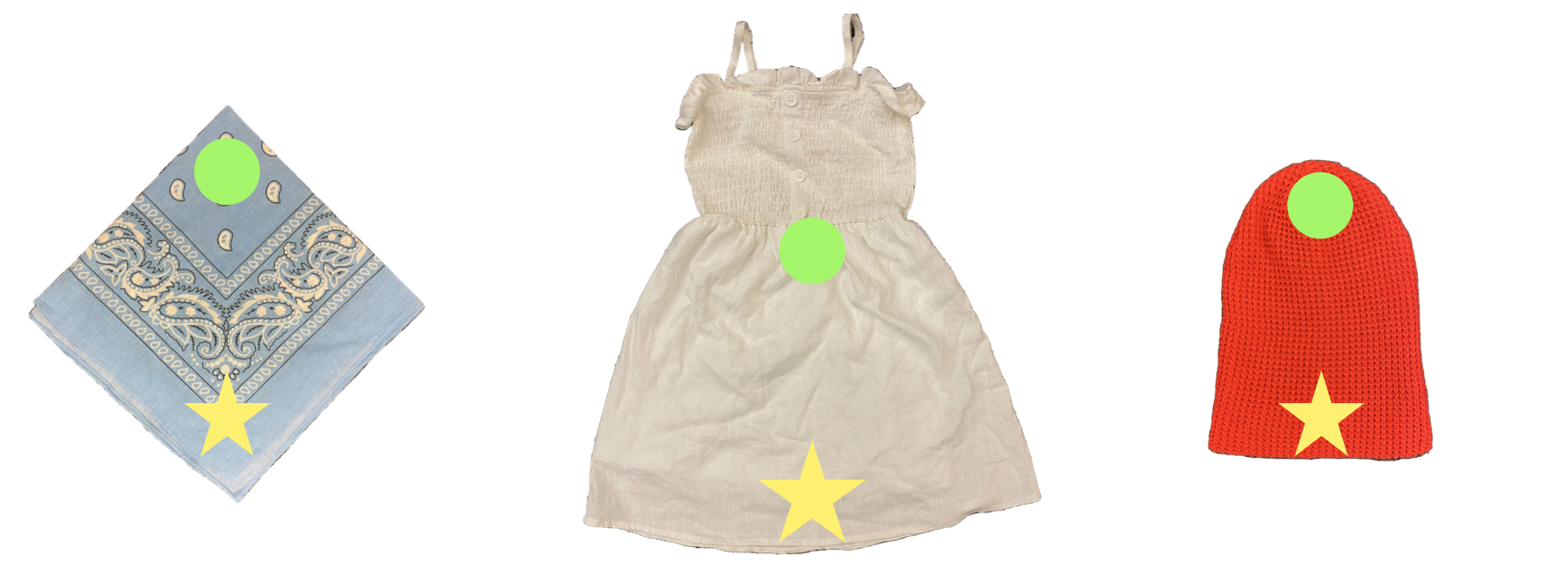}
\caption{
Non-bag experiments. Left to right: Folded cloth, dress, hat. The robot pins at the region indicated by the green point and attempts to grasp a single layer at the yellow star region.
}
\vspace{-10pt}
\label{fig:garments}
\end{figure}

\begin{table}[t]
  \setlength\tabcolsep{5.0pt}
  \centering
  \vspace{-0em}
  \vspace{1pt}
  %\begin{adjustbox}{width=\columnwidth,center}
  \centering
  \footnotesize
  \begin{tabular}{c |ccc| c}
  \toprule
  \multirow{2}{*}{Objects}  &  \multicolumn{3}{c}{0-Shot Recall} &   \multicolumn{1}{|c}{\multirow{2}{*}{SLIP Success Rate}}  \\
  \cmidrule(lr){2-4}
   & 0-layer & 1-layer & 2-layer & \\
  \midrule
  Folded Cloth & 100\% & 100\%  & 62\% & 4/6 \\
  Dress & 100\% & 75\%  & 75\% & 4/6\\
  Hat & 100\% & 83\%  & 25\% & 5/6\\
  \bottomrule
  \end{tabular}
  \caption{Non-bag experiments. 
  The middle 3 columns show the (multi-class) recall of the video classification model trained on bags and tested on garments without finetuning. The last column shows the success rate of grasping a single layer using \slip with the classification model.
  }
  \label{tab:garment_result}
  \vspace*{-10pt}
\end{table}

\subsection{Bagging Experiments Results}\label{subsec:bag_results}

Figure~\ref{fig:grasp_height} shows the distribution of the number of layers grasped at various grasp heights (measured from the surface height) for each of the 4 training bags in the training data. As expected, as the grasp height decreases, it is less likely to grasp 0 layers and more likely to grasp 2 layers. However, while some grasp heights are more likely than others to grasp a single layer for each bag, there is no single grasp height that always works, as the success of single-layer grasping depends highly on the specific configuration of the bags. 

Results in Table~\ref{tab:bag_result} demonstrate that \bagging achieves a higher success rate than all baselines for all bags. In most trials, \bagging takes fewer than 15 actions to flatten the bag and \slip successfully singulates a layer within 4 grasp iterations. Among the baselines, Perceived-Depth has a low success rate of grasping a single layer. This is because, for the mesh bag, the perceived depth is often too deep due to holes, resulting in 2-layer grasps, while for other bags, the perceived depth is often not deep enough and leads to 0-layer grasps. For the Handle Grasping baseline, it is not applicable to drawstring bags since they do not have handles, and while it achieves a high success rate on handbags, it is not effective for plastic bags. The handles of plastic bags are on the two sides, so grasping and lifting them does not help create an opening like with a handbag. Its failures on handbags are mainly due to mistakenly grasping the handle associated with the bottom layer or accidentally grasping both handles. For AutoBag, which is designed for thin plastic bags, is not effective on drawstring bags and fabric handbags. This is because its key action for opening the bag, ``Compress,'' uses air to inflate the bag during a downward motion, and so only works for bags with lightweight material and without holes.

We observe 4 failure modes of \bagging:
\begin{enumerate}[(A)]
    \item Failure to flatten the bag with the correct orientation.
    \item Failure to successfully grasp a single layer of the bag.
    \item Bag slips out of the gripper after grasping a single layer.
    \item Robot hand hits the bag handles during insertion and does not put the objects inside the bag.
\end{enumerate}

Failure (A) occurs when the perception module fails to recognize the rim and handle regions, and thus rotates the bag in the wrong orientation. Failure (B) 
%occurs sometimes due to wrong predictions by the video classification model in \slip, 
usually occurs when the robot accidentally grasps both layers and manipulates the bag into a configuration 
%, causing the layers to roll into the bottom during the movement due to friction with the workspace surface. This can 
which prevents the robot from grasping a single layer in future trajectories. For bags with stiff plastic and fabric materials, failure (C) can occur when the single-layer grasp is not firm enough and the bag slips out. % gripper during further actions such as rotation and insertion (when the other hand accidentally hits the bag, it may knock it off). 
Failure (D) is often due to the bag not being rotated completely sideways, which means the insertion hand can fail to enter the bag. %Additionally, for thin and thick plastic bags, their opening size is relatively small compared to their handles. Thus, during insertion, 
Additionally, the robot hand may sometimes hit the handles or other parts of the bag, pushing the bag away or causing it to rotate, leading to insertion failure.

\subsection{Single-Layer Grasping on Fabrics}

We test \slip on other materials to evaluate its applicability to general single-layer grasping tasks. We consider 3 deformable objects: a blue piece of cloth folded twice into a square (Fig.~\ref{fig:teaser}), a white dress, and a red hat (Fig.~\ref{fig:garments}). The task goal is to grasp their top layer only. We apply our video classification model to these objects without any finetuning. 

Table~\ref{tab:garment_result} shows the 0-shot multi-class recall metrics for the classification model as well as the success rate of achieving a single-layer grasp. In each case, the model predicts accurately on a 0-layer grasp and 1-layer grasp, but less accurately on a 2-layer grasp, for which there are greater visual differences across objects.  
%We hypothesize that this is because the visual appearances of a 0-layer and 1-layer are similar across objects, while they appear different under 2-layer grasps. 
A failure mode associated with grasping a folded cloth is that the cloth has 4 layers. Grasping 1, 2, and 3 layers look visually similar, so the model would mistaken those 2- and 3-layer grasps as a 1-layer grasp. 
While the model accuracy is lower than that of bags the model is trained on, the SLIP success rate is fairly high. This is because if the robot starts from a grasp higher than the surface and gradually decreases its height, it suffices for the model to accurately recognize a 1-layer grasp.

Out of the 5 failed trials, 3 are due to model prediction errors, such as  
%incorrectly classifying a 0-layer or 2-layer grasp as a 1-layer grasp when it is not, or 
misclassifying a 1-layer (or 2-layer) grasp and adjusting the grasp height in the wrong direction. The other 2 failures occur from the object slipping out of the robot's 1-layer grasp during its gripper movement. % When a 1-layer grasp is weak, it may slip as the robot pulls it backward and forward. Unlike elastic plastic bags that have the tendency to recover to their original state, fabrics remain in the middle of the trajectory where it slips out of the gripper. Thus, when the robot tries to regrasp at the original position, the top layer is no longer there to be grasped.

\section{Conclusion and Future Work}

In this paper, we propose an approach for bagging by singulating layers using interactive perception. %Experiments show that \bagging achieves significantly higher success rates over baselines for opening a bag, inserting items, and lifting the bag. In future work, we plan to apply this approach to related tasks such as packing and wrapping. 
Future directions include speeding up the pipeline, applying the approach to related tasks such as packing and wrapping, and studying scenarios where two sides of the bag stick together tightly.

%We hope that this work will lead to an exciting era in robotic manipulation of bags and 3D deformable objects more generally.

% Daniel: can remove if needed.
\section*{Acknowledgments}
{\footnotesize
This research was performed at the AUTOLAB at UC Berkeley in affiliation with the Berkeley AI Research (BAIR) Lab, and the CITRIS ``People and Robots'' (CPAR) Initiative. The authors are supported in part by donations from Toyota Research Institute and equipment grants from NVIDIA.  L.Y. Chen is supported by the National Science Foundation (NSF) Graduate Research Fellowship Program under Grant No. 2146752. D. Seita and D. Held are supported by NSF CAREER grant IIS-2046491. We thank Ryan Burgert for providing rubber ducks for our experiments and Kaushik Shivakumar for valuable feedback.
}

% Daniel: this is the old way of doing it. The alternative with \printbibliography I think will be better since we can customize it a little more and we can add hyperref (which didn't work when I tried this variant).
% \clearpage
% \balance  % We need this line for the correct ordering of reference. ( this line is in "balance" Package
%{\footnotesize
%\bibliographystyle{IEEEtran}
%\bibliography{references}
%}

% Daniel: these two commands need to be paired up with the \addbibresource{references.bib} command at the top where `references.bib` is the name of our file.
\renewcommand*{\bibfont}{\footnotesize}
\printbibliography

% \clearpage
% \input{appendix}

\end{document}